%% file: neurips_2026.tex
\documentclass{article}

\usepackage[preprint]{neurips_2026}

\usepackage{graphicx}
\usepackage{amsmath}
\usepackage{amsthm}
\usepackage{amssymb,amsfonts}
\usepackage{booktabs}

\usepackage{makecell}

\usepackage{dsfont}

\usepackage{comment}
\usepackage{amsmath}

\usepackage{multirow}

\newcommand{\imineq}[2]{\includegraphics[width=#2cm]{#1}}

\usepackage{threeparttable}

\usepackage{enumitem}

\usepackage{bbding}

\usepackage{natbib}
\setcitestyle{numbers,square}
\usepackage{wrapfig}
\usepackage{gensymb}
\usepackage{xparse} 
\usepackage{caption}
\usepackage{multicol}
\usepackage{subcaption}
\usepackage{caption}
\usepackage[utf8]{inputenc} % allow utf-8 input
\usepackage[T1]{fontenc}    % use 8-bit T1 fonts
\usepackage{hyperref}       % hyperlinks
\usepackage{url}            % simple URL typesetting
\usepackage{booktabs}       % professional-quality tables
\usepackage{amsfonts}       % blackboard math symbols
\usepackage{nicefrac}       % compact symbols for 1/2, etc.
\usepackage{microtype}      % microtypography
\usepackage{xcolor}         % colors
\usepackage{algorithm}
\usepackage{algorithmic}
\usepackage{multirow}
\usepackage{booktabs}
\usepackage{colortbl} 
\usepackage{xcolor}
\usepackage{array}  
\usepackage{booktabs}

\usepackage{physics}

\usepackage{xspace}

\usepackage[utf8]{inputenc} % allow utf-8 input
\usepackage[T1]{fontenc}    % use 8-bit T1 fonts
\usepackage{hyperref}       % hyperlinks
\usepackage{url}            % simple URL typesetting
\usepackage{booktabs}       % professional-quality tables
\usepackage{amsfonts}       % blackboard math symbols
\usepackage{nicefrac}       % compact symbols for 1/2, etc.
\usepackage{microtype}      % microtypography
\usepackage{xcolor}         % colors

\usepackage{microtype}
\usepackage{caption}
\usepackage{graphicx}
\usepackage{booktabs} % for professional tables
\usepackage{graphicx}
\usepackage{amsmath}
\usepackage{hyperref}
\usepackage{multirow}
\usepackage{subcaption}
\usepackage{textcomp}
\usepackage{booktabs}
\usepackage{colortbl}  %彩色表格需要加载的宏包
\usepackage{xcolor}
\usepackage{array}  
\usepackage{booktabs}
\usepackage[table]{xcolor}
\usepackage{colortbl}

\usepackage{pifont}
\newcommand{\cmark}{\textcolor{green!60!black}{\ding{51}}} % ✓
\newcommand{\xmark}{\textcolor{red!70!black}{\ding{55}}}   % ✗ 

\title{Learning to Align Generative Appearance Priors for Fine-grained Image Retrieval}
\author{
Shijie Wang$^{1}$, Yadan Luo$^{1}$, Zijian Wang$^{1}$, Xin Yu$^{2}$, Zi Huang$^{1}$ \\
$^{1}$ The University of Queensland, Australia \quad
$^{2}$ The University of Adelaide, Australia
}

\begin{document}

\maketitle

\begin{abstract}
Fine-grained image retrieval (FGIR) typically relies on supervision from seen categories to learn discriminative embeddings for retrieving unseen categories. However, such supervision often biases retrieval models toward the semantics of seen categories rather than the underlying appearance characteristics that generalize across categories, thereby limiting retrieval performance on unseen categories.
To tackle this, we propose GAPan, a \underline{\textbf{G}}enerative \underline{\textbf{A}}ppearance \underline{\textbf{P}}rior \underline{\textbf{a}}lignment \underline{\textbf{n}}etwork that reformulates the learning objective from category prediction toward appearance modeling.
Technically, GAPan treats retrieval features with an invertible density model based on normalizing flows. 
In the \textit{forward }direction, the flow maps all instance features into a latent density space, where each seen category is modeled by a class-conditional Gaussian prior and optimized via exact likelihood estimation. This formulation preserves richer appearance details by leveraging the invertible property of the flows.
In the \textit{reverse} direction, samples from the high-density regions of these learned priors are mapped back to the feature space to produce appearance-aware anchors that reflect intra-category variation. These anchors supervise a prior-driven alignment objective that aligns retrieval embeddings with category-specific appearance distributions, thereby improving generalization to unseen categories.
Evaluations demonstrate that our GAPan achieves state-of-the-art performance on both widely-used fine- and coarse-grained benchmarks.
\end{abstract}

\section{Introduction}
Fine-grained image retrieval (FGIR) aims to retrieve images from the same fine-grained subcategory within a common meta-category. It supports applications with large and evolving visual taxonomies, from fine-grained clothing retrieval~\cite{DBLP:conf/cvpr/AkKLT18,DBLP:conf/cvpr/LiuLQWT16} to ecological monitoring~\cite{DBLP:conf/cvpr/ElhoseinyZZE17,wei2021fine}. The difficulty is that the relevant evidence is often subtle and local, while the space of subcategories is naturally open-ended. A retrieval model must therefore go beyond separating the labels observed during training and preserve appearance factors that remain useful for novel subcategories.

Most existing FGIR methods \cite{DBLP:conf/iccv/KoGK21,DBLP:conf/icml/RothMOCG21, DBLP:journals/corr/abs-2512-06255,DBLP:conf/iccv/ZhengZL021} address this challenge through label-driven discriminative learning. Category labels are commonly used to construct discriminative objectives, such as instance pairs, tuples, proxies, or semantic regularizers \cite{DBLP:conf/iclr/Furusawa24,DBLP:conf/eccv/KirchhofRAK22,DBLP:conf/eccv/TehDT, DBLP:journals/pr/YanXSLS24}. These objectives are effective because they directly shape the similarity structure used for retrieval. However, they use labels mainly to specify which training instances should be close or far apart, rather than to model how appearance varies within a fine-grained category. This becomes limiting when the test subcategories differ from the training subcategories. A model may rely on cues that separate seen classes but do not transfer to unseen ones. For example, if training bird species are largely separated by head color, the embedding may overemphasize crown color, whereas novel species with similar heads may require wing patterns or beak shape for retrieval. Thus, category supervision should not only define discriminative relations, but also capture the \textit{distribution} of appearance variations underlying fine-grained similarity.

\begin{wrapfigure}{r}{0.6\textwidth}
		%\vspace{-4mm}
		\small 
		%\centering\vspace{-3ex}
  \includegraphics[width=1\linewidth]{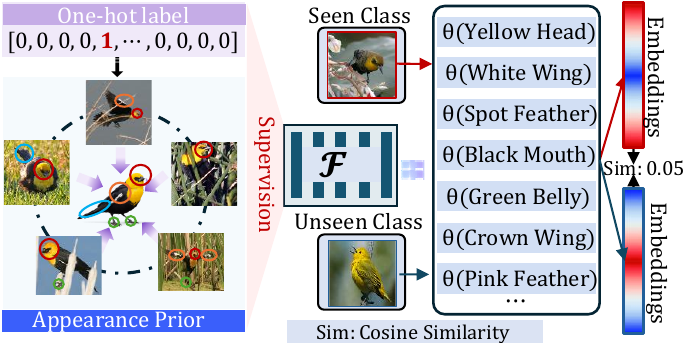}
  \caption{ Motivation of GAPan. Beyond one-hot labels, GAPan aggregates diverse instance-level cues into category-specific appearance priors, providing appearance-aware supervision that preserves visual properties and improves generalization to unseen categories.  }
  \vspace{-3ex}
  \label{introd}
	\end{wrapfigure}
    
As illustrated in Fig.~\ref{introd}, we therefore reinterpret category labels as cues for aggregating instance observations into distribution-level appearance priors. A label alone only specifies category membership; it does not describe which visual cues are reliable or how they vary across instances. We address this by modeling the retrieval features of each category as a class-conditional appearance distribution, rather than a single prototype. This provides a complementary supervisory signal: contrastive learning establishes a retrieval-oriented feature manifold, while appearance priors regularize this manifold with category-level distributional structure.

To instantiate this idea, we introduce GAPan, a Generative Appearance Prior alignment network based on normalizing flows (NF)~\cite{lei2023pyramidflow,mate2022flowification}. GAPan treats retrieval features with an invertible density model:  $\imineq{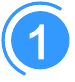}{0.3}$  In the \textit{forward} direction, the flow maps retrieval features to a latent density space, where each seen category is modeled by a class-conditional Gaussian prior and optimized by exact likelihood. Because the NF mapping is invertible, this density modeling does not require compressing fine-grained information into a fixed prototype. $\imineq{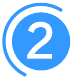}{0.3}$  In the \textit{reverse} direction, samples from high-density regions of the learned priors are mapped back to the retrieval feature space, producing appearance-aware anchors that reflect category-level variation. These anchors are used in a prior-guided alignment objective, which pulls each instance toward anchors from its own category while contrasting it against anchors from other categories. GAPan is trained progressively: the retrieval backbone is first warmed up with an auxiliary contrastive objective to establish a stable retrieval manifold, and is then jointly optimized with likelihood modeling, posterior calibration, and prior-guided alignment. This joint objective converts category labels from discrete indicators of membership into distribution-level supervision over the retrieval space. Compared with single-prototype or purely discriminative supervision, GAPan preserves intra-category appearance variation while encouraging category-level consistency, leading to representations that generalize better to novel fine-grained subcategories.

Our main contributions are summarized below:

\begin{itemize}
    \item  To the best of our knowledge, we are the first to shift supervisory signals from category prediction toward appearance modeling, thereby improving generalization to unseen categories.

    \item We propose GAPan, an invertible flow framework that distills diverse retrieval features into explicit category-specific appearance priors and leverages samples drawn from their high-density regions as generative supervisory signals for appearance-aware alignment.

    \item Extensive experiments demonstrate that GAPan achieves state-of-the-art performance on widely used fine-grained and coarse-grained image retrieval benchmarks.

\end{itemize}
\section{Related Work}
\noindent\textbf{Fine-grained Image Retrieval.} Recent advances in fine-grained image retrieval primarily follow two technical paradigms~\cite{DBLP:conf/ijcai/ZhengJSWHY18,DBLP:conf/wacv/MoskvyakMDB21, DBLP:conf/iccv/KoGK21,DBLP:conf/icml/RothMOCG21,DBLP:conf/iccv/ZhengZL021,DBLP:conf/cvpr/KimKCK21,DBLP:conf/cvpr/ZhengWL021}. Localization-based methods, exemplified by A$^2$-Net~\cite{wei20212}, focus on precise object or part localization to enhance retrieval accuracy via reconstruction-based learning. In parallel, metric-based approaches such as DDML~\cite{DBLP:conf/aaai/ParkPNK25} learn discriminative embedding spaces through tailored distance metrics, while NIA~\cite{DBLP:conf/cvpr/RothVA22} enforces unique translatability from class proxies, encouraging same-subcategory samples to cluster in the embedding space.
Despite their effectiveness, these methods mainly derive supervision from category labels and discriminative relations, which provides limited guidance for explicitly modeling the appearance distributions behind fine-grained categories. As a result, subtle visual patterns that are shared or comparable across categories may not be sufficiently captured. In contrast, GAPan learns category-specific appearance priors from feature distributions within a shared latent space. By leveraging these priors to generate appearance-aware anchors for supervision, GAPan encourages embeddings to capture fine-grained appearance cues while maintaining discrimination for retrieving both seen and unseen categories.

\noindent\textbf{Normalizing Flow.} 
Normalizing flows \cite{hirschorn2023normalizing,wang2022low} are a class of powerful generative models that transform complex data distributions into simple, tractable ones via invertible and differentiable mappings. They have attracted considerable attention due to their flexibility and precise likelihood estimation in high-dimensional settings. For instance, Glow~\cite{kingma2018glow} employs invertible $1 \times 1$ convolutions and exact log-likelihood optimization for high-quality image generation. Beyond generative modeling, normalizing flows also demonstrate strong performance in classification tasks. STG-NF ~\cite{lei2023pyramidflow}  learns the data distribution and scores the samples according to their likelihood, while TWGC~\cite{mackowiak2021generative} highlights their robustness and interpretability compared to standard feed-forward models. Moreover, Flow-OOD~\cite{kirichenko2020normalizing} demonstrates their effectiveness in out-of-distribution detection.
Unlike these works, GAPan leverages normalizing flows to model explicit category-specific appearance priors from instance features and further exploits the inverse flow to generate appearance-aware anchors for distribution-level supervision, complementing cross-category discrimination.

\begin{figure*}[t]
\begin{center}
   \includegraphics[width=1\linewidth]{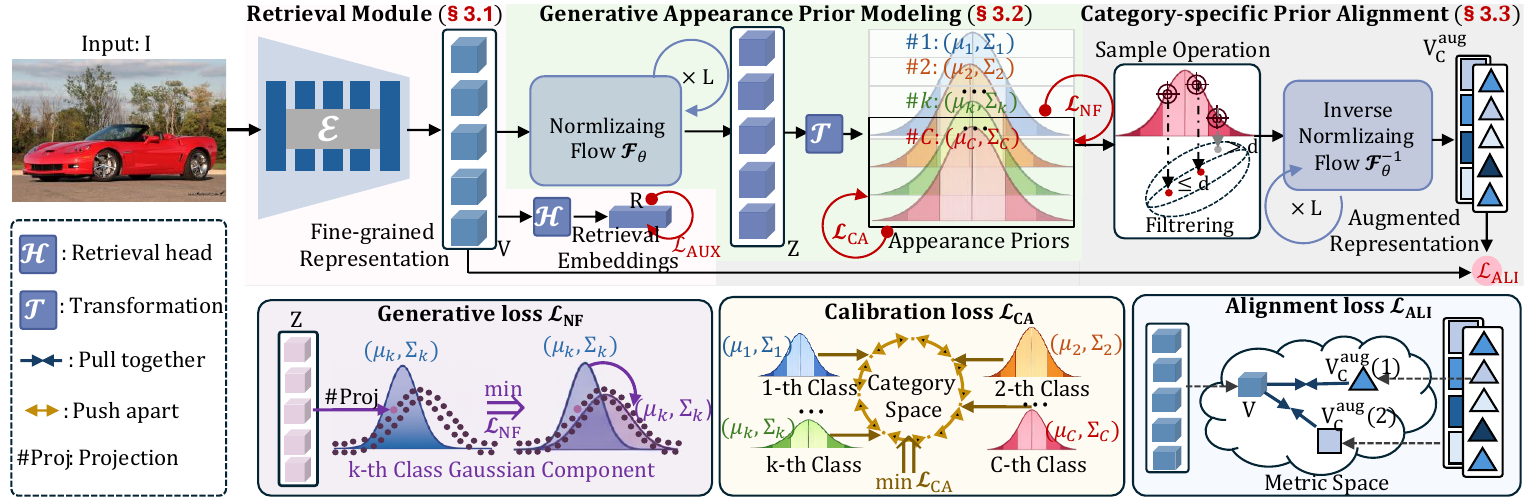}
\end{center}
\setlength{\abovecaptionskip}{-0.1cm} 
   \caption{Detailed illustration of {
   \bf Generative Appearance Prior alignment network}. See \S\ref{GAPan} for more details. }
\label{approach}
\end{figure*}
\section{Method}
\label{GAPan}
The core of GAPan (Fig.~\ref{approach}) lies in shifting the supervisory paradigm from category prediction to appearance modeling. The framework integrates a retrieval backbone with two flow-driven modules: 
\begin{itemize}
    \item Forward Flow: Generative Appearance Prior Modeling maps instance features into a unified latent manifold, where its invertible property distills diverse observations into explicit category-specific Gaussian distributions while retaining all visual details.

    \item Reverse Flow: Category-specific Prior Alignment samples from high-density regions of these priors to generate appearance-aware anchors that provide distribution-level supervision, guiding retrieval embeddings toward underlying category-specific appearance characteristics.
\end{itemize}

\subsection{Retrieval Module}
The retrieval module is responsible for extracting visual representations and mapping them into a discriminative embedding space. Specifically, given an input image $\mathrm{I\in \mathbb{R}^{3\times H \times W}}$, the retrieval network $\mathcal{E}(\cdot)$ first extracts a high-dimensional feature embedding $\mathrm{V  = \mathcal{E}(I) \in \mathbb{R}^{C}}$. 
To ensure a fair comparison with existing works~\cite{DBLP:conf/aaai/ParkPNK25,DBLP:conf/iclr/Ren0WH24}, we use a lightweight retrieval head $\mathcal{H}(\cdot)$ to project $\mathrm{V}$ into a compact $\mathrm{C'}$-dimensional embedding space, yielding the final retrieval embedding $\mathrm{R=\mathcal{H}(\mathrm{V})\in \mathbb{R}^{C'}}$, where $\mathrm{C'=128}$.
Note that GAPan introduces no additional computational overhead at inference.

\subsection{Generative Appearance Prior Modeling}
To address the limited capacity of label-driven supervision to capture cross-category appearance comparability, we propose a generative appearance prior modeling module that learns category-specific priors in a shared latent space. This module utilizes \textit{the forward transformation of normalizing flows} to map instance features into a unified latent manifold. Its invertibility ensures an information-preserving transformation, distilling diverse observations into explicit category-specific Gaussian distributions without losing the subtle details typical of discriminative compression.

\noindent\textbf{Feature Mapping via Normalizing Flows.}
To explicitly model appearance priors, we employ an invertible normalizing flow $\mathcal{F}_\theta(\cdot)$ to map instance features $\mathrm{V}$ into a latent space $\mathrm{Z}$. This lossless transformation captures complex distributions while preserving fine-grained information.
Specifically, we adopt a modified RealNVP architecture~\cite{DBLP:conf/iclr/DinhSB17}, where the transformation $\mathrm{Z}=\mathcal{F}_\theta(\mathrm{V})$ is achieved through $\mathrm{L}$ iterative coupling layers. Each layer partitions the input into $[\mathrm{Z^{(l-1)}_1,\mathrm{Z}^{(l-1)}_2}]$ and applies an affine transformation to one partition conditioned on the other:

\begin{equation}
\mathrm{Z^{(l)}_1 = \mathrm{Z}^{(l-1)}_1, \quad
\mathrm{Z}^{(l)}_2 =
\mathrm{Z}^{(l-1)}_2 \odot \exp\left(s^{(l)}(\mathrm{Z}^{(l-1)}_1)\right)
+
t^{(l)}(\mathrm{Z}^{(l-1)}_1)},
\end{equation}
where $\mathrm{Z^{(0)}=\mathrm{V}}$ and $\mathrm{Z^{(L)}=\mathrm{Z}=\mathcal{F}_{\theta}(\mathrm{V})}$. Here, $\mathrm{s^{(l)}(\cdot)}$ and $\mathrm{t^{(l)}(\cdot)}$ denote the scale and translation networks in the $\mathrm{l}$-th coupling layer, respectively, and both are implemented as MLPs.
By stacking coupling layers with alternating partitions, the flow progressively maps complex feature distributions into a tractable latent space, where appearance priors can be effectively modeled.

\noindent\textbf{Generative Probabilistic Modeling.}
%After mapping instance features into the latent space, we model category-specific appearance priors with class-conditional Gaussian distributions. 
Within the unified latent manifold established by the forward flow, we characterize category-specific appearance priors using class-conditional Gaussian distributions.
Each category $\mathrm{k}$ is assigned a Gaussian distribution:
\begin{equation}
\mathrm{P_Z(Z \mid k) = \mathcal{N}(Z \mid \mu_k, \Sigma_k)},
\end{equation}
where $\mathrm{\mu}_k\in \mathbb{R}^{C}$ and $\mathrm{\Sigma}_k\in \mathbb{R}^{C}$ denote the mean vector and the diagonal covariance matrix of the Gaussian distribution for the $k$-th category, respectively. This formulation represents the appearance prior of each category as a Gaussian distribution that captures rich intra-category variations.

Using the change-of-variables formula, we connect the latent appearance prior to the observed feature space. For an instance feature $\mathrm{V}$ with category $\mathrm{y}$, its class-conditional density is given by
\begin{equation}
\mathrm{P_V(V \mid k=y, \theta) = \mathcal{N}(\mathcal{F}_\theta(V) \mid \mu_k, \Sigma_k) \cdot \left| \det \frac{\partial \mathcal{F}_\theta(V)}{\partial V} \right| }.
\label{eq3}
\end{equation}
By coupling the flow transformation with the class-conditional Gaussian distributions, the model forms a unified probabilistic framework that organizes instance-level variations into category-specific appearance priors for fine-grained representation learning.

\noindent\textbf{Normalizing Flow Optimization.} 
To improve the accuracy of density modeling and discriminative ability of the learned appearance priors, a dual-objective function is employed. First, the generative loss $\mathcal{L}_{\mathrm{NF}}$ is formulated as the negative log-likelihood (NLL) under Eq.~\eqref{eq3}:
\begin{equation}
\mathrm{ \mathcal{L}_{\mathrm{NF}} = -\frac{1}{|\mathcal{B}|} \sum_{(V_i, y_i) \in \mathcal{B}} \log \mathrm{P_V(V_i \mid k=y_i, \theta)}},
\label{eq:nll}
\end{equation}
where $\mathcal{B}$ and $|\mathcal{B}|$ denote a mini-batch and its corresponding size, respectively.
Minimizing the NLL jointly optimizes the flow parameters $\theta$ and Gaussian parameters $\mathrm{\{\mu_k,\Sigma_k\}_{k=1}^K}$, encouraging each instance feature to lie in a high-density region of its class-conditional distribution, where $\mathrm{K}$ denotes the number of seen categories.

To preserve discriminative ability, we introduce a posterior calibration constraint $\mathcal{L}_{\mathrm{CA}}$ that encourages category distributions to remain separable. The posterior probability of an instance belonging to class $k$ is computed by normalizing its likelihood across all class-conditional distributions:
\begin{equation}
\mathrm{P(k \mid V) = \frac{\mathcal{N}(\mathcal{F}_\theta(V) \mid \mu_k, \Sigma_k)}{\sum_{j=1}^{K} \mathcal{N}(\mathcal{F}_\theta(V) \mid \mu_j, \Sigma_j)}}.
\label{eq:gmm_posterior}
\end{equation}
$\mathcal{L}_{\mathrm{CA}}$ minimizes the cross-entropy between this posterior and the corresponding ground-truth label $\mathrm{y_i}$:
\begin{equation}
\mathrm{\mathcal{L}_{\mathrm{CA}} = -\frac{1}{|\mathcal{B}|} \sum_{(V_i, y_i) \in \mathcal{B}} \log \mathrm{P(k=y_i \mid V_i)}}.
\label{eq:ca_loss}
\end{equation}
While the NLL increases the likelihood under the target category distribution, the posterior calibration loss normalizes the likelihood across categories and explicitly promotes inter-category separation. Together, these objectives couple the flow transformation with the class-conditional Gaussian distributions, producing appearance priors that faithfully characterize category-specific visual patterns and remain discriminative for fine-grained retrieval.

\subsection{Category-specific Prior Alignment}
To directly transfer appearance priors into the retrieval embedding space, we introduce a category-specific prior alignment module that exploits \textit{the inverse property of normalizing flows}, sampling from high-density regions of the learned Gaussian priors to generate appearance-aware anchors in the original feature space. These anchors serve as distribution-level supervisory signals, guiding the retrieval model to align instance embeddings with the underlying appearance characteristics of their corresponding categories.

For each class $\mathrm{k}$, this module samples $\mathrm{N}_{\mathrm{S}}$ latent vectors $\mathrm{S_{k,i} \in \mathbb{R}^{C}}$ from the corresponding Gaussian component $\mathcal{N}(\mathrm{\mu}_{\mathrm{k}}, \mathrm{\Sigma}_{\mathrm{k}})$ via rejection sampling. These latent vectors are then transformed back into the feature space via the inverse flow $\mathcal{F}_{\theta}^{-1}(\cdot)$, producing synthesized feature anchors $\mathrm{V^{\mathrm{aug}}_{k,i} \in \mathbb{R}^{C}}$:
\begin{equation}
\mathrm{V^{\mathrm{aug}}_{k,i} = \mathcal{F}_{\theta}^{-1}(\mathrm{S}_{k,i}), \quad 
\mathrm{S}_{k,i} \sim \mathcal{N}(\mathrm{\mu}_{\mathrm{k}}, \mathrm{\Sigma}_{\mathrm{k}}), \quad 
\text{s.t.} \quad
\sqrt{(\mathrm{S}_{k,i} - \mathrm{\mu}_{\mathrm{k}})^\top \mathrm{\Sigma}_{\mathrm{k}}^{-1}(\mathrm{S}_{k,i} - \mathrm{\mu}_{\mathrm{k}})} \leq \mathrm{d}}.
\label{eq:sampling}
\end{equation}
where $\mathrm{d}>0$ is a threshold controlling the sampling region within the Gaussian distribution.
The threshold $\mathrm{d}$ governs the trade-off between sampling reliability and diversity: a smaller $\mathrm{d}$ restricts sampling to high-probability regions, yielding typical feature anchors that capture reliable category-specific appearance cues, whereas a larger $\mathrm{d}$ explores a wider region to provide more diverse anchors.

For the $\mathrm{m}$-th training instance with label $\mathrm{y_m}$, generated features from the same category, \textit{i.e.}, $\{\mathrm{V^{\mathrm{aug}}_{y_m,i}\}_{i=1}^{N_S}}$, are treated as positive anchors, while those from other categories serve as negatives. These generated features are projected into the retrieval space by the same head $\mathcal{H}(\cdot)$ used for real instance features, so that each retrieval embedding $\mathrm{R_m}$ can be aligned with its category-matched synthesized anchors via the prior-guided alignment loss:
\begin{equation}
\mathrm{\mathcal{L}_{\mathrm{ALI}}
=
-\frac{1}{|\mathcal{B}| \times N_S}
\sum_{m=1}^{|\mathcal{B}|}
\sum_{i=1}^{N_S}
\log
\frac{
\exp(\mathrm{sim}(R_m, \mathcal{H}(V^{aug}_{y_m,i})) / \tau)
}{
\sum_{c=1}^{K}
\sum_{j=1}^{N_S}
\exp(\mathrm{sim}(R_m, \mathcal{H}(V^{aug}_{c,j})) / \tau)
}},
\label{ali}
\end{equation}
where $\mathrm{sim}(\cdot,\cdot)$ denotes cosine similarity and $\tau$ is a temperature parameter. This loss injects learned appearance priors into the retrieval embedding space, encouraging representations to be organized around appearance cues rather than category semantics. Consequently, the alignment objective promotes capturing transferable appearance cues, which improves generalization to unseen categories.

\subsection{Overall Training Objective}
To ensure stable convergence and generative-discriminative alignment, GAPan adopts a progressive two-stage training protocol separated by a threshold epoch $\mathrm{T}$.

\noindent\textbf{Auxiliary Contrastive Learning.} 
During the initial stage, the retrieval model is preconditioned using an auxiliary contrastive loss $\mathcal{L}_{\mathrm{AUX}}$ to establish a robust feature manifold. For a mini-batch $\mathcal{B}$ containing $\mathrm{N}$ categories with two instances per category, the batch size is $\mathrm{|\mathcal{B}|=2N}$. For each instance $\mathrm{i}$, let $p(i)$ denote the index of the other instance of the same category in $\mathcal{B}$ (with $p(i)\neq i$). The loss is defined as:
\begin{equation}
\mathrm{\mathcal{L}_{\mathrm{AUX}} = - \frac{1}{|\mathcal{B}|} \sum_{i=1}^{|\mathcal{B}|} \log \frac{\exp(\mathrm{sim}(R_i, R_{p(i)}) / \tau)}{ \sum_{m=1, m \neq i}^{|\mathcal{B}|} \exp(\mathrm{sim}(R_i, R_m) / \tau)}},
\label{aux}
\end{equation}
where $\tau$ is the same temperature as in Eq.~\eqref{ali}.

%, $\tau$ is a temperature hyperparameter. %The distance function $\mathrm{D}(\cdot,\cdot)$ is implemented as the squared Euclidean distance between $\ell_2$-normalized vectors $\mathrm{D}(\mathrm{R_i}, \mathrm{R_j}) = \left| \frac{\mathrm{R_i}}{\left|\mathrm{R_i}\right|_2} - \frac{\mathrm{R_j}}{\left|\mathrm{R_j}\right|_2} \right|^2_2 = 2 - 2 \frac{\langle\mathrm{R_i}, \mathrm{R_j}\rangle}{\left|\mathrm{R_i}\right|_2 \cdot \left|\mathrm{R_j}\right|_2}$.

\noindent\textbf{Progressive Training Schedule.} The optimization process is partitioned into a warmup phase and a joint optimization phase based on the current epoch $\mathrm{t}$: 
\begin{itemize}
\item \textbf{Phase I: Warmup ($\mathrm{t \leq T}$):} 
During the warmup phase, only the retrieval model is optimized with the auxiliary contrastive loss, providing stable feature representations for subsequent flow-based prior modeling. The total objective is $\mathrm{\mathcal{L} = \mathcal{L}_{\mathrm{AUX}}}$.
\item 
\textbf{Phase II: Joint Optimization ($\mathrm{t > T}$):} After epoch $\mathrm{T}$, the normalizing flow and the retrieval model are jointly optimized by combining contrastive supervision, probabilistic prior modeling, calibration, and prior-based alignment:
\begin{equation}
    \mathrm{\mathcal{L}_{total} = \mathcal{L}_{\mathrm{AUX}} + \alpha \mathcal{L}_{\mathrm{NF}} + \beta \mathcal{L}_{\mathrm{CA}} + \gamma \mathcal{L}_{\mathrm{ALI}}},
    \label{total}
\end{equation}
where $\alpha$, $\beta$, and $\gamma$ are hyperparameters that balance the contributions of the corresponding loss terms.
\end{itemize}

\section{Experiments}
\subsection{Experimental Setup}
\textbf{Datasets.} CUB-200-2011 \cite{Branson2014Bird} consists of 200 bird species. We use the first 100 categories (5,864 images) for training and the remaining 100 categories (5,924 images) for testing. 
Stanford Cars \cite{Krause20133D} includes 196 car models. Similarly, we use the first 98 categories (8,054 images) for training and the remaining 98 categories (8,131 images) for testing. 
Stanford Online Products (SOP) \cite{DBLP:conf/cvpr/SchroffKP15} is divided into 11,318 categories (59,551 images) for training and 11,316 categories (60,502 images) for testing. 
\textit{This split ensures \textbf{no category overlap} between training and testing sets, where all testing categories are strictly unseen during training to evaluate cross-category generalization.}

\noindent\textbf{Implementation Details.}
Our retrieval model is implemented on top of ViT-B/16~\cite{dosovitskiy2020image} pre-trained on ImageNet-21K \cite{DBLP:conf/nips/RidnikBNZ21}. Input images are resized to $256 \times 256$ and randomly cropped to $224 \times 224$ during training. We adopt stochastic gradient descent with an initial learning rate of $1 \times 10^{-5}$, momentum of 0.9, and weight decay of $1 \times 10^{-4}$, using a batch size of 270 on two NVIDIA A6000 GPUs. The learning rate follows an exponential decay schedule with a decay factor of 0.9 every five epochs over a total of 50 training epochs. The scale and translation networks of normalizing flow are multi-layer perceptrons with a hidden dimension of 512. $\tau$ is set to 0.09 in Eqs.~\eqref{ali} and \eqref{aux}.

\noindent\textbf{Evaluation protocol.} We evaluate retrieval performance using Recall@K with cosine distance, following the standard protocol in prior work \cite{DBLP:conf/cvpr/SongXJS16}. Specifically, for each query image, the model retrieves the top-$\mathrm{K}$ most similar images. A score of 1 is assigned if at least one positive image appears within the top-$\mathrm{K}$ results, and 0 otherwise. The final Recall@K is obtained by averaging these scores across all queries in the test set.

\subsection{Ablation Experiments}

\begin{wraptable}{c}{0.45\textwidth}
		\vspace{-1.3em}
		 \footnotesize 
        \centering
	\caption{ Comparison on CUB-200-2011 dataset using different combinations of constraints. 
	}
	\begin{tabular}{cccc|c}
	  
		\toprule[1pt]
		 $\mathcal{L}_{\mathrm{AUX}}$&$\mathcal{L}_{\mathrm{NF}}$&$\mathcal{L}_{\mathrm{CA}}$ &$\mathcal{L}_{\mathrm{ALI}}$ &Recall@1\\
		\toprule[0.7pt]
		\cmark & \xmark & \xmark & \xmark & 82.6\% \\
  \midrule
        \cmark & \cmark & \xmark & \xmark & 84.7\% \\
        \cmark & \cmark & \cmark & \xmark & 85.3\% \\
        \cmark & \cmark & \cmark & \cmark & 89.1\% \\

		\toprule[1pt]
	\end{tabular}\\
	\label{ablation}
		\vspace{-1.0em}
	\end{wraptable}
We evaluate the contribution of each component in GAPan on the CUB-200-2011 dataset. As shown in Tab. \ref{ablation}, $\mathcal{L}_{\mathrm{AUX}}$ provides the basic retrieval supervision and establishes the embedding space for appearance prior modeling. Adding $\mathcal{L}_{\mathrm{NF}}$ improves Recall@1 from 82.6\% to 84.7\%, showing that coupling the flow transformation with class-conditional Gaussian distributions helps organize instance-level variations into category-specific appearance priors. With $\mathcal{L}_{\mathrm{CA}}$, the performance further increases to 85.3\%, indicating that posterior calibration preserves discriminative ability by encouraging better separation among category-specific Gaussian distributions. The largest gain comes from $\mathcal{L}_{\mathrm{ALI}}$, which improves Recall@1 by 3.8\%. This loss samples feature anchors from high-density regions of the learned category distributions and uses them as additional positive anchors in a contrastive objective. Instead of directly forcing each instance to match a single prototype, $\mathcal{L}_{\mathrm{ALI}}$ aligns instance embeddings with category-specific appearance priors while maintaining separation from other categories. This appearance-prior-guided supervision reduces reliance on sparse category labels and leads to stronger generalization for fine-grained retrieval.

\subsection{Comparison with the State-of-the-Art Methods}
\noindent\textbf{Fine-grained Image Retrieval.}
As shown in Tab. \ref{t2}, GAPan achieves state-of-the-art performance on two widely-used fine-grained retrieval benchmarks, CUB-200-2011 and Stanford Cars 196. It obtains Recall@1 scores of 89.1\% and 92.6\%, outperforming the previous best method, LaFG \cite{DBLP:journals/corr/abs-2512-06255}, by absolute margins of 1.9\% and 1.1\%, respectively. Existing competitive methods, such as DDML \cite{DBLP:conf/aaai/ParkPNK25} and SEE \cite{DBLP:conf/ijcai/LeW25}, mainly rely on category-level supervision to learn discriminative embeddings, which may limit their ability to capture visual regularities that transfer to unseen categories. In contrast, GAPan distills diverse instance observations into explicit appearance priors, enabling the model to learn visual patterns beyond category labels. Its improvement over LaFG, which incorporates external VLM knowledge, further demonstrates the effectiveness of appearance-prior-guided alignment learned directly from instance distributions.

\noindent\textbf{Generalization on Coarse-grained Retrieval.}
%To further evaluate the scalability and generalization capacity of GAPan, we conduct evaluations on the large-scale Stanford Online Products (SOP) dataset. Even in this coarse-grained context, GAPan maintains its lead with a Recall@1 of 88.4\%, surpassing the latest competitive models such as DPHM \cite{DBLP:journals/pr/XuCH25} (+3.6\%) and LaFG (+1.3\%) \cite{DBLP:journals/corr/abs-2512-06255}. Typical FGIR models often struggle on SOP due to the lack of fine-grained subcategory names or the high variance within coarse labels. However, by leveraging category labels as navigational indicators to aggregate instance-level information, GAPan successfully distills reliable appearance priors that are robust to semantic scale. The consistent improvements across both fine-grained and coarse-grained domains substantiate that grounding representation learning in generative visual consistency—rather than discrete label-fitting—effectively enhances cross-category comparability and ensures more discriminative and transferable embeddings.
To further evaluate the generalization capacity of GAPan, we conduct experiments on the large-scale Stanford Online Products (SOP) dataset. Even in this coarse-grained setting, GAPan achieves a Recall@1 of 88.4\%, surpassing recent competitive methods such as DPHM \cite{DBLP:journals/pr/XuCH25} (+3.6\%) and LaFG \cite{DBLP:journals/corr/abs-2512-06255} (+1.3\%). SOP poses a different challenge from fine-grained benchmarks, as its coarse labels contain larger intra-class variation and provide weaker fine-grained semantic cues. By using category labels to organize instance-level distributions, GAPan distills reliable appearance priors that remain effective across different levels of category granularity. The consistent improvements across both fine-grained and coarse-grained datasets show that appearance-prior-guided supervision, rather than relying only on discrete category labels, improves cross-category comparability and yields more discriminative and generalizable embeddings.

\begin{table*}[!t]\centering
	\caption{ Compared with competitive methods on CUB-200-2011, Stanford Cars 196 and Stanford Online Products datasets. ``Arch'' denotes the backbone architecture. ``R50'' and ``ViT'' denote the ResNet-50~\cite{He2015Deep} and Vision Transformer~\cite{dosovitskiy2020image} backbones, respectively.
} 
\resizebox{\textwidth}{!}{
	\begin{tabular}{l|l|cccc|cccc|cccc}
		\toprule[1pt]
       \multirow{2.5}{*}{Method} & \multirow{2.5}{*}{Arch}& \multicolumn{4}{c|}{CUB-200-2011} &\multicolumn{4}{c|}{Stanford Cars 196} & \multicolumn{4}{c}{Stanford Online Products}\\
		\cline{3-14}
		 &  & 1&2&4&8& 1&2&4&8& 1&10&100&1000 \\
         \toprule[0.7pt]
         CBML$_{\rm TPAMI23}$ \cite{DBLP:journals/pami/KanHCLMH23}&R50& 69.9 &80.4 &87.2 &92.5& 88.1 & 92.6 &95.4&97.4 &79.9 & 91.5&  96.5&98.9\\  
         NIR$_{\rm CVPR22}$ \cite{DBLP:conf/cvpr/RothVA22} &R50& 70.5 & 80.6 & -&-&89.1 & 93.4 &-&-&80.4 & 91.4& -& - \\
          HSE$_{\rm ICCV23}$ \cite{DBLP:conf/iccv/YangSLCCS23}& R50 & 70.6& 80.1& 87.1 & -&89.6& 93.8& 96.0& -&80.0 &91.4&96.3& - \\
          
IDML$_{\rm TPAMI24}$ \cite{DBLP:journals/pami/WangZZZL24} & R50 & 70.7& 80.2& - & -&90.6& 94.5& -&  -& 81.5&-& -& - \\
HIST $_{\rm  CVPR{22}}$ \cite{DBLP:conf/cvpr/LimYP022} & R50& 71.4 &  81.1& 88.1&-&89.6& 93.9& 96.4&-&81.4 & 92.0& 96.7& - \\
PNCA++$_{\rm ECCV20}$ \cite{DBLP:conf/eccv/TehDT}& R50& 72.2 & 82.0 & 89.2 &93.5& 90.1& 94.5&97.0&98.4&81.4&  92.4& 96.9&99.0\\
			\toprule[0.7pt]
DIML$_{\rm TPAMI24}$ \cite{DBLP:journals/pami/ZhaoRZL24} & ViT & 76.7 & -&-&-& 80.7 &-&-&-& 79.5 &-&-&- \\
%LM-Metric$_{\rm PR24}$ \cite{DBLP:journals/pr/YanXSLS24} & ViT & 77.1  & -&-&-& - &-&-&-& 83.1 &-&-&- \\
DFML-PA$_{\rm CVPR23}$ \cite{DBLP:conf/cvpr/WangZL0L23} & ViT & 79.1& 86.8& -& -& 89.5& 93.9& - &- &84.2& 93.8& -& -\\
DVA $_{\rm AAAI26}$ \cite{jiang2025fine} & ViT & 84.9 &90.6 &94.5 &96.7& 90.7 &94.8& 97.1& 98.4& -& -& -& -\\
DPHM$_{\rm PR25}$ \cite{DBLP:journals/pr/XuCH25} & ViT& 85.5 & 91.3 & 94.6 & 96.6 & 84.1 & 90.5 & 94.5 & 97.1 & 84.8 & 94.5 & 98.1 & 99.4 \\
HypViT$_{\rm CVPR22}$ \cite{DBLP:conf/cvpr/ErmolovMKSO22} & ViT & 85.6 & 91.4 & 94.8& 96.7& 89.2 & 94.1 & 96.7 &98.1&85.9&94.9&98.1&99.5\\
HIER$_{\rm CVPR23}$ \cite{DBLP:conf/cvpr/KimJK23} & ViT & 85.7 & 91.3 & 94.4 & - & 88.3 & 93.2 & 96.1 & - & 86.1&95.0&98.0&-\\
SEE$_{\rm IJCAI25}$ \cite{DBLP:conf/ijcai/LeW25} &ViT &85.8 &91.4 &94.6& -& 88.8 &93.8 &96.4 &-&86.3 &95.0 &98.2& -\\
DDML$_{\rm AAAI25}$ \cite{DBLP:conf/aaai/ParkPNK25} & ViT & 86.0 & 91.7 & 95.2 & 96.8 & 89.5 & 94.2 & 96.8 & 98.2 & 86.1&95.1&98.2&99.5\\
VPTSP-G$_{\rm ICLR24}$ \cite{DBLP:conf/iclr/Ren0WH24} & ViT & 86.6 & 91.7 & 94.8 & - & 87.7 & 93.3 & 96.1 & - & 84.4&93.6&97.3&-\\
 LaFG$_{\rm ARXIV25}$ \cite{DBLP:journals/corr/abs-2512-06255} & ViT & 87.2 & 92.4 & 95.2 & 97.0 & 91.5 & 94.6& 96.6 & 98.5 & 87.1 &95.8& 98.5 & 99.5 \\
\toprule[0.7pt]
\rowcolor{gray!30}
Our GAPan & ViT & \textbf{89.1} & \textbf{93.5}&\textbf{95.3}&\textbf{97.1}& \textbf{92.6} & \textbf{95.7} & \textbf{97.6} & \textbf{98.7} & \textbf{88.4}& \textbf{96.6}&\textbf{98.9} & \textbf{99.6} \\

\toprule[1pt]
	\end{tabular}}
    \label{t2}
\end{table*} 

\subsection{Further Analysis}

\begin{wraptable}{c}{0.55\textwidth}
		 \vspace{-1.3em}
		 \footnotesize 
        \centering
\caption{Retrieval accuracy on CUB-200-2011 of model trained with different number of flow layers $\mathrm{L}$.}
\begin{tabular}{c|ccccc}
\toprule[1pt]
Number $\mathrm{L}$ &1& 2 & 4 & 8 & 16 \\
\toprule[0.7pt]
Recall@1 &84.7\% & 86.3\% & 87.9\% & 89.1\% & 87.8\% \\
\toprule[1pt]
\end{tabular}
		\vspace{-1.8em}
        \label{layer}
	\end{wraptable}

\textbf{Effect on the depth of normalizing flow.} 
We further investigate the sensitivity of GAPan to the depth of the normalizing flow by varying the number of coupling layers $\mathrm{L}$ from 1 to 16. As summarized in Tab. \ref{layer}, Recall@1 on CUB-200-2011 first increases with the flow depth and reaches the best performance of 89.1\% at $\mathrm{L}=8$, then drops to 87.8\% when $\mathrm{L}=16$. When the flow is too shallow, its limited transformation capacity makes it difficult to map complex instance features into a structured latent space for appearance prior modeling. In contrast, an overly deep flow introduces more learnable parameters and may overfit the relatively limited fine-grained training data. These results suggest that an appropriate flow depth allows the model to capture complex feature distributions without introducing unnecessary optimization difficulty.

\begin{wraptable}{c}{0.55\textwidth}
		 \vspace{-1.3em}
		 \footnotesize 
        \centering
\caption{Accuracy on CUB-200-2011 for models trained with different numbers of sampling anchors.}
\begin{tabular}{c|ccccc}
\toprule[1pt]
Number $\mathrm{N_S}$ &1& 2 & 4 & 6 & 8 \\
\toprule[0.7pt]
Recall@1 &87.6\% & 88.0\% & 88.4\% & 89.1\% & 87.8\% \\
\toprule[1pt]
\end{tabular}
		\vspace{-1.8em}
        \label{sam}
	\end{wraptable}

\textbf{Sensitivity on the number of synthesized anchors.} 
As shown in Tab. \ref{sam}, we evaluate the effect of the number of generated anchors $\mathrm{N_S}$ in GAPan. Recall@1 increases from 87.6\% to 89.1\% as $\mathrm{N_S}$ grows from 1 to 6, showing that more anchors provide richer supervision from the learned category-specific appearance priors. However, further increasing $\mathrm{N_S}$ to 8 reduces performance to 87.8\%. This suggests that excessive generated anchors may introduce redundant constraints and weaken the preservation of instance-level variations. These results suggest that an appropriate number of generated anchors strengthens prior-guided supervision without suppressing instance-level variation.

\textbf{Impact of sampling threshold in Eq.~\eqref{eq:sampling}.} 
As shown in Tab. \ref{thresh}, we evaluate the effect of the sampling threshold d in Eq. \eqref{eq:sampling}. Recall@1 reaches the best performance of 89.1\% at $\mathrm{d}=1.0$. When $\mathrm{d}$ is too small, such as $\mathrm{d}=0.2$, sampling is restricted to a narrow high-density region around the center of the 
\begin{wraptable}{c}{0.55\textwidth}
		 \vspace{-0.1em}
		 \footnotesize 
        \centering
\caption{Accuracy on CUB-200-2011 for models trained with diverse threshold $\mathrm{d}$ for generated anchors.}
\begin{tabular}{c|ccccc}
\toprule[1pt]
Threshold $\mathrm{d}$ &0.2& 0.5 & 1 & 1.5 & 2.0 \\
\toprule[0.7pt]
Recall@1 &86.1\% & 87.9\% & 89.1\% & 88.3\% & 87.4\% \\
\toprule[1pt]
\end{tabular}
		\vspace{-1.8em}
     \label{thresh}
	\end{wraptable}
Gaussian distribution, producing reliable but less diverse anchors and thus limiting the supervision provided by the learned appearance priors. When $\mathrm{d}$ is too large, such as $\mathrm{d}=2.0$, sampling covers a broader region of the Gaussian distribution and may include less representative anchors, which can introduce noisy supervision. The optimal threshold $\mathrm{d}=1.0$ provides a better balance between anchor reliability and diversity, leading to stronger fine-grained retrieval performance.

\begin{wraptable}{c}{0.55\textwidth}
		 \vspace{-1.3em}
		 \footnotesize 
        \centering
\caption{Retrieval accuracy on CUB-200-2011 with different warm-up epochs $\mathrm{T}$.}
\begin{tabular}{c|ccccc}
\toprule[1pt]
Epoch $\mathrm{T}$ &0& 2 & 5 & 8 & 10 \\
\toprule[0.7pt]
Recall@1 &88.2\% & 88.8\% & 89.1\% & 89.0\% & 88.7\% \\
\toprule[1pt]
\end{tabular}
		\vspace{-1.8em}
        \label{warmup}
	\end{wraptable}

\textbf{Effect of warm-up epochs.} 
We examine the impact of the warm-up strategy by varying the number of initial warm-up epochs $\mathrm{T}$. As shown in Tab. \ref{warmup}, Recall@1 reaches the best performance of 89.1\% at $\mathrm{T}=5$. Without warm-up, the normalizing flow is optimized on unstable early-stage features, which weakens appearance prior modeling. When $\mathrm{T}$ is too small, the embedding space is not sufficiently stable for the flow to estimate reliable category distributions. In contrast, an overly long warm-up period leaves fewer epochs for optimizing the flow and the appearance-prior-guided supervision, leading to slightly lower performance. These results suggest that a moderate warm-up period provides sufficiently stable features for appearance prior modeling while leaving enough training time for global optimization.

\begin{wrapfigure}{r}{0.4\textwidth}
		\vspace{-1.0em}
		\small 
		\centering
  \includegraphics[width=1\linewidth]{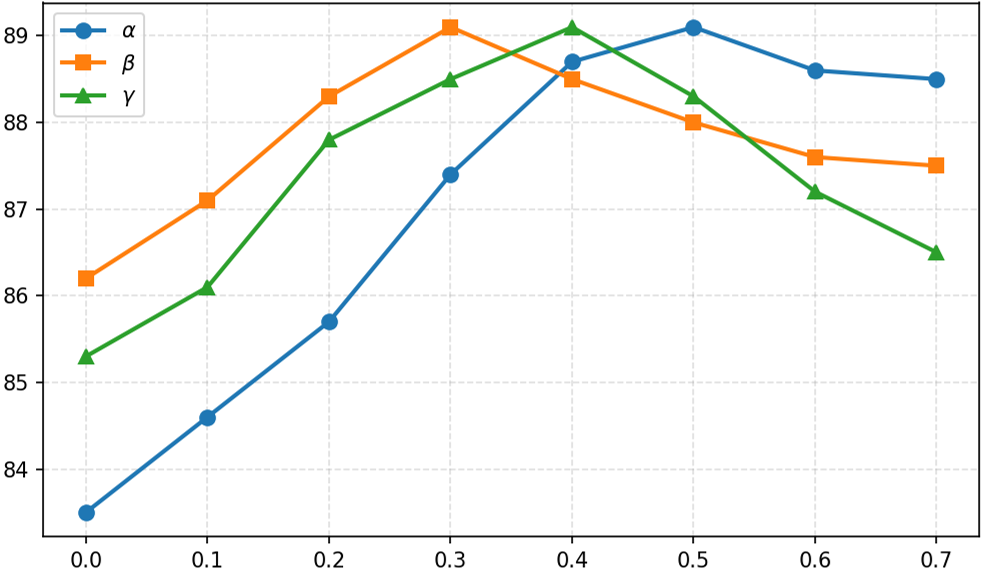}

  \caption{ Hyperparameter analysis of $\alpha$, $\beta$ and $\gamma$ in Eq.~\eqref{total}.
  }
  \label{hyper}
\vspace{-1em}
	\end{wrapfigure}

\textbf{Hyperparameter analysis.}
We conduct sensitivity analyses of the hyperparameters in Eq. \eqref{total}, with the results shown in Fig. \ref{hyper}. The performance of GAPan varies with the weights $\alpha, \beta$, and $\gamma$, showing that these objectives need to be properly balanced. When these weights are too small, the corresponding objectives, including distribution modeling, posterior calibration, and feature sampling contrastive supervision, contribute insufficiently to training. In contrast, overly large weights may overemphasize the auxiliary objectives and weaken retrieval discrimination. Based on these observations, we therefore set $\alpha = 0.5$, $\beta = 0.3$, and $\gamma = 0.4$ as the default values.

\begin{wraptable}{c}{0.4\textwidth}
		 \vspace{-1.3em}
		 \footnotesize 
        \centering
\caption{Performance comparison between GAPan and point-based representation strategies, including K-means clustering and prototypical learning, on CUB-200-2011.}
\begin{tabular}{c|ccc}
\toprule[1pt]
Method & K-means & Proto & GAPan  \\
\toprule[0.7pt]
Recall@1 &83.9\% & 85.4\% &89.1\%  \\
\toprule[1pt]
\end{tabular}
		\vspace{-1.8em}
        \label{point}
	\end{wraptable}
\textbf{Effectiveness of generative appearance priors.} 
As shown in Tab. \ref{point}, we compare GAPan with point-based representation strategies, including K-means clustering and prototypical learning, to evaluate the effectiveness of generative appearance priors. K-means represents feature structures with discrete centroids, which may smooth out fine-grained variations within each category. Prototypical learning uses a fixed prototype for each category, making it difficult to describe diverse intra-category appearances. In contrast, GAPan models each category as a probabilistic distribution in a flow-based latent space and generates multiple appearance anchors from high-density regions of the learned distributions. This distribution-based formulation provides richer supervision than point-based representations, leading to better cross-category comparability and stronger generalization.

\begin{wraptable}{c}{0.4\textwidth}
		 \vspace{-1.3em}
		 \footnotesize 
        \centering
\caption{Retrieval accuracy on CUB-200-2011 for models trained with either a normalizing flow (NF) or an encoder–decoder (ED) architecture implemented as a multi-layer MLP.}
\begin{tabular}{c|cc}
\toprule[1pt]
Method & ED & NF ($\mathcal{F}_\theta$)  \\
\toprule[0.7pt]
Recall@1 &84.9\% &89.1\%  \\
\toprule[1pt]
\end{tabular}
		\vspace{-1.8em}
        \label{NF}
	\end{wraptable}

\textbf{Importance of normalizing flows.} 
We compare GAPan with an encoder-decoder (ED) baseline implemented as a multi-layer MLP, using a comparable parameter budget, to evaluate the contribution of the flow-based architecture to appearance prior modeling. %We keep the same losses ($\mathcal{L}_{\mathrm{NF}}, \mathcal{L}_{\mathrm{CA}}, \mathcal{L}_{\mathrm{ALI}}$) and training schedule unchanged.
As shown in Tab. \ref{NF}, replacing the normalizing flow with ED reduces Recall@1 from 89.1\% to 84.9\%. This indicates that a standard non-invertible mapping is less effective in connecting the observed feature space with the latent category distributions. In contrast, the normalizing flow provides an invertible transformation and enables exact likelihood estimation through the change-of-variables formula, allowing instance features and class-conditional Gaussian distributions to be coupled in a unified probabilistic framework. This coupling keeps the sampled anchors consistent with the learned category distributions, providing more effective appearance-prior-guided supervision for FGIR.

\begin{figure}[!t]
\centering
\begin{tabular}{cc}
\begin{minipage}[t]{0.48\linewidth}

    \includegraphics[width = 1\linewidth]{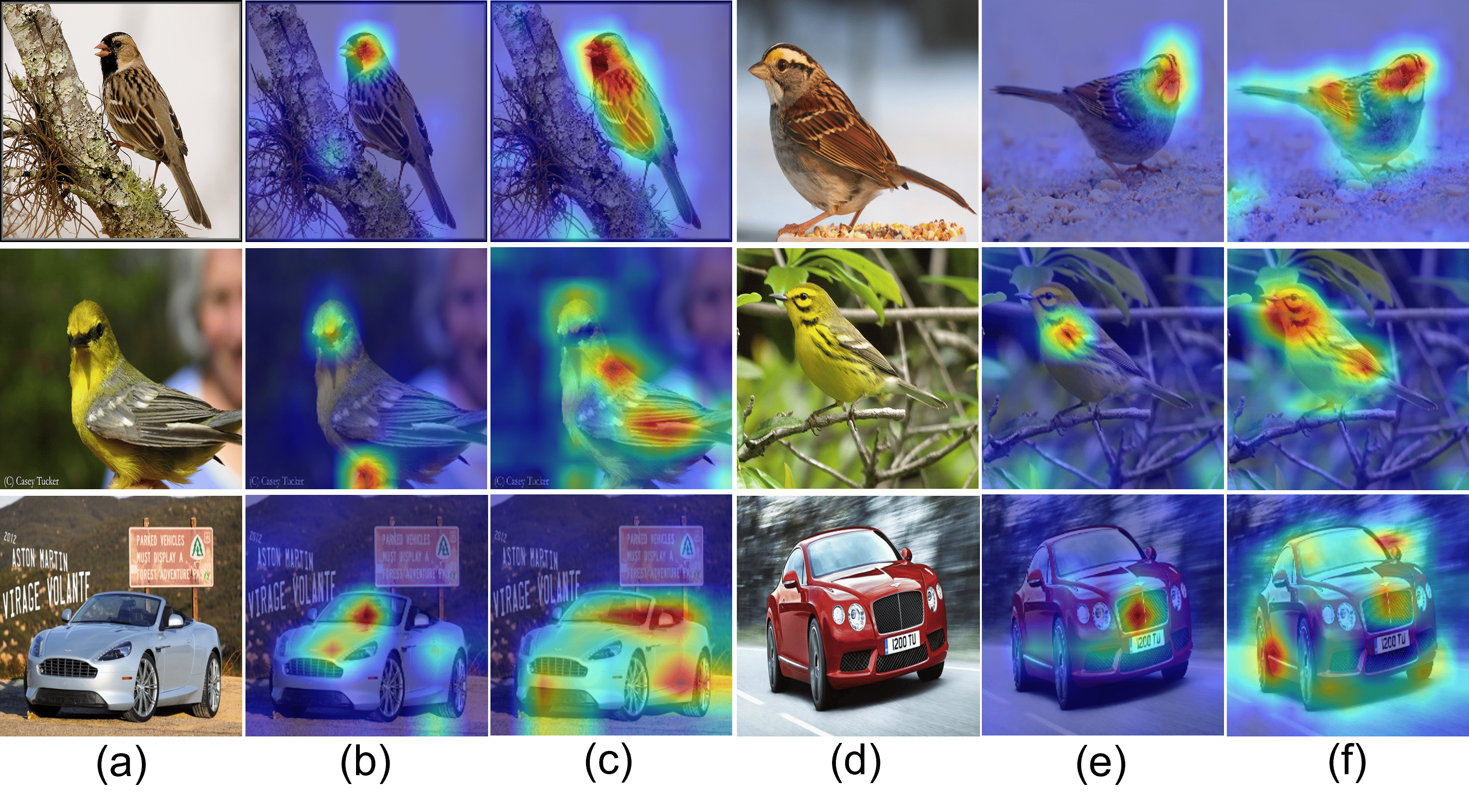}
    \caption{Illustration of class activation maps. (a) and (d) denote the input images, (b) and (e) show the activation maps produced by the baseline, and (c) and (f) present those generated by GAPan.}
    \label{feature}
\end{minipage}
&
\begin{minipage}[t]{0.48\linewidth}
    \includegraphics[width = 1\linewidth]{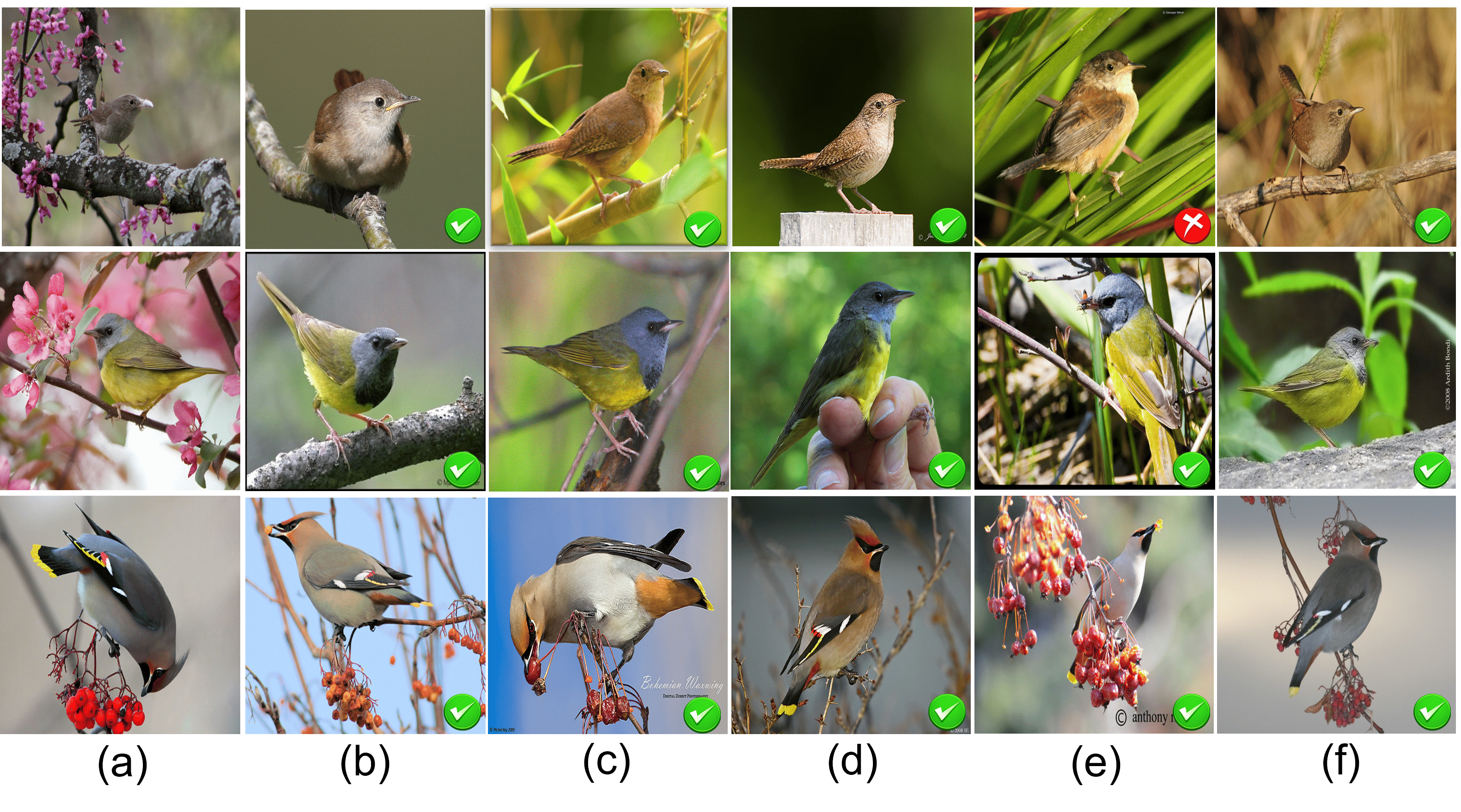}
        \caption{Examples of top-5 retrieval results on CUB-200-2011 by GAPan. (a) denotes the query image, and (b)--(f) show the retrieved images in descending order of similarity.}
    \label{result}
\end{minipage}
\end{tabular}
\end{figure}

\textbf{Feature Representation Analysis.} Since the learned appearance priors are optimized in a latent space and cannot be directly visualized, we interpret their effect indirectly using class activation maps and retrieval examples, which respectively show the attended visual regions and the resulting cross-instance retrieval behavior. As shown in Fig.~\ref{feature}, the baseline mainly activates the most discriminative local regions, which is consistent with its reliance on category-level supervision. In contrast, GAPan produces broader and more complete activation responses over object regions, covering additional visual cues such as body parts, texture, and contour. This indicates that the proposed appearance priors encourage the model to capture category-specific appearance patterns beyond isolated discriminative regions. Consistent with this broader visual perception, the retrieval examples in Fig.~\ref{result} show that GAPan retrieves images with similar fine-grained appearance despite variations in pose and background. Given a query image, GAPan retrieves visually similar birds with consistent fine-grained characteristics, even when pose, scale, or background varies. This suggests that the learned embeddings are guided by transferable appearance cues rather than only by category-specific decision boundaries. Together, the activation maps and retrieval results show that GAPan improves representation learning by aligning instance embeddings with learned appearance priors, leading to more complete object perception and stronger generalization to unseen fine-grained categories.

\section{Conclusion and Discussion}
In this paper, we introduced GAPan, a framework that leverages the invertible nature of normalizing flows to establish a shared latent density space, shifting the learning paradigm from category prediction toward appearance modeling. Through the forward flow, GAPan models diverse instance observations with explicit category-specific Gaussian priors, while the reverse flow generates appearance-aware anchors to provide distribution-level supervision. This bidirectional generative framework effectively aligns retrieval embeddings with transferable appearance characteristics across categories. Extensive evaluations demonstrate that GAPan consistently outperforms existing methods and exhibits superior generalization to novel subcategories across both fine- and coarse-grained benchmarks.

\textbf{Broader Impact.}
Fine-grained retrieval is pivotal for specialized domains such as biodiversity monitoring, precision agriculture, and e-commerce. GAPan advances FGIR by shifting from sparse label-based supervision to distribution-aware appearance modeling. This paradigm is particularly valuable for low-resource or open-set scenarios where new or rare categories frequently emerge; by capturing underlying visual priors, our model reduces the dependency on exhaustive manual annotations and improves generalization to unseen classes. This capability supports more robust and scalable visual analysis in real-world applications where data labeling is often a bottleneck.

\textbf{Limitations.} The deployment of FGIR systems requires careful validation of data integrity. When the training distribution is long-tailed or lacks sufficient visual diversity, the learned appearance priors tend to inherit these biases, leading to degraded and inconsistent performance across domains. In real-world applications, such as endangered species monitoring or commercial recommendation, we advocate using GAPan as a decision-support tool rather than an autonomous system, with human-in-the-loop verification to mitigate risks arising from distribution shifts and domain gaps.

{
\small
\bibliographystyle{plain}
\bibliography{ref}
}

%%%%%%%%%%%%%%%%%%%%%%%%%%%%%%%%%%%%%%%%%%%%%%%%%%%%%%%%%%%%

%%%%%%%%%%%%%%%%%%%%%%%%%%%%%%%%%%%%%%%%%%%%%%%%%%%%%%%%%%%%

\newpage
\input{checklist.tex}

\end{document}

%% file: checklist.tex
\section*{NeurIPS Paper Checklist}

\begin{enumerate}

\item {\bf Claims}
    \item[] Question: Do the main claims made in the abstract and introduction accurately reflect the paper's contributions and scope?
    \item[] Answer: \answerYes{} % Replace by \answerYes{}, \answerNo{}, or \answerNA{}.
    \item[] Justification: We make the main claims clearly in the abstract and introduction.
    \item[] Guidelines:
    \begin{itemize}
        \item The answer \answerNA{} means that the abstract and introduction do not include the claims made in the paper.
        \item The abstract and/or introduction should clearly state the claims made, including the contributions made in the paper and important assumptions and limitations. A \answerNo{} or \answerNA{} answer to this question will not be perceived well by the reviewers. 
        \item The claims made should match theoretical and experimental results, and reflect how much the results can be expected to generalize to other settings. 
        \item It is fine to include aspirational goals as motivation as long as it is clear that these goals are not attained by the paper. 
    \end{itemize}

\item {\bf Limitations}
    \item[] Question: Does the paper discuss the limitations of the work performed by the authors?
    \item[] Answer: \answerYes{} % Replace by \answerYes{}, \answerNo{}, or \answerNA{}.
    \item[] Justification: We discuss the limitation of this paper in Section 5.
    \item[] Guidelines:
    \begin{itemize}
        \item The answer \answerNA{} means that the paper has no limitation while the answer \answerNo{} means that the paper has limitations, but those are not discussed in the paper. 
        \item The authors are encouraged to create a separate ``Limitations'' section in their paper.
        \item The paper should point out any strong assumptions and how robust the results are to violations of these assumptions (e.g., independence assumptions, noiseless settings, model well-specification, asymptotic approximations only holding locally). The authors should reflect on how these assumptions might be violated in practice and what the implications would be.
        \item The authors should reflect on the scope of the claims made, e.g., if the approach was only tested on a few datasets or with a few runs. In general, empirical results often depend on implicit assumptions, which should be articulated.
        \item The authors should reflect on the factors that influence the performance of the approach. For example, a facial recognition algorithm may perform poorly when image resolution is low or images are taken in low lighting. Or a speech-to-text system might not be used reliably to provide closed captions for online lectures because it fails to handle technical jargon.
        \item The authors should discuss the computational efficiency of the proposed algorithms and how they scale with dataset size.
        \item If applicable, the authors should discuss possible limitations of their approach to address problems of privacy and fairness.
        \item While the authors might fear that complete honesty about limitations might be used by reviewers as grounds for rejection, a worse outcome might be that reviewers discover limitations that aren't acknowledged in the paper. The authors should use their best judgment and recognize that individual actions in favor of transparency play an important role in developing norms that preserve the integrity of the community. Reviewers will be specifically instructed to not penalize honesty concerning limitations.
    \end{itemize}

\item {\bf Theory assumptions and proofs}
    \item[] Question: For each theoretical result, does the paper provide the full set of assumptions and a complete (and correct) proof?
    \item[] Answer: \answerNA{} % Replace by \answerYes{}, \answerNo{}, or \answerNA{}.
    \item[] Justification: This paper does not include theoretical results.
    \item[] Guidelines:
    \begin{itemize}
        \item The answer \answerNA{} means that the paper does not include theoretical results. 
        \item All the theorems, formulas, and proofs in the paper should be numbered and cross-referenced.
        \item All assumptions should be clearly stated or referenced in the statement of any theorems.
        \item The proofs can either appear in the main paper or the supplemental material, but if they appear in the supplemental material, the authors are encouraged to provide a short proof sketch to provide intuition. 
        \item Inversely, any informal proof provided in the core of the paper should be complemented by formal proofs provided in appendix or supplemental material.
        \item Theorems and Lemmas that the proof relies upon should be properly referenced. 
    \end{itemize}

    \item {\bf Experimental result reproducibility}
    \item[] Question: Does the paper fully disclose all the information needed to reproduce the main experimental results of the paper to the extent that it affects the main claims and/or conclusions of the paper (regardless of whether the code and data are provided or not)?
    \item[] Answer:\answerYes{}
    \item[] Justification: We discuss the experimental setup in Section 4.
    \item[] Guidelines:
    \begin{itemize}
        \item The answer \answerNA{} means that the paper does not include experiments.
        \item If the paper includes experiments, a \answerNo{} answer to this question will not be perceived well by the reviewers: Making the paper reproducible is important, regardless of whether the code and data are provided or not.
        \item If the contribution is a dataset and\slash or model, the authors should describe the steps taken to make their results reproducible or verifiable. 
        \item Depending on the contribution, reproducibility can be accomplished in various ways. For example, if the contribution is a novel architecture, describing the architecture fully might suffice, or if the contribution is a specific model and empirical evaluation, it may be necessary to either make it possible for others to replicate the model with the same dataset, or provide access to the model. In general. releasing code and data is often one good way to accomplish this, but reproducibility can also be provided via detailed instructions for how to replicate the results, access to a hosted model (e.g., in the case of a large language model), releasing of a model checkpoint, or other means that are appropriate to the research performed.
        \item While NeurIPS does not require releasing code, the conference does require all submissions to provide some reasonable avenue for reproducibility, which may depend on the nature of the contribution. For example
        \begin{enumerate}
            \item If the contribution is primarily a new algorithm, the paper should make it clear how to reproduce that algorithm.
            \item If the contribution is primarily a new model architecture, the paper should describe the architecture clearly and fully.
            \item If the contribution is a new model (e.g., a large language model), then there should either be a way to access this model for reproducing the results or a way to reproduce the model (e.g., with an open-source dataset or instructions for how to construct the dataset).
            \item We recognize that reproducibility may be tricky in some cases, in which case authors are welcome to describe the particular way they provide for reproducibility. In the case of closed-source models, it may be that access to the model is limited in some way (e.g., to registered users), but it should be possible for other researchers to have some path to reproducing or verifying the results.
        \end{enumerate}
    \end{itemize}

\item {\bf Open access to data and code}
    \item[] Question: Does the paper provide open access to the data and code, with sufficient instructions to faithfully reproduce the main experimental results, as described in supplemental material?
    \item[] Answer: \answerNo{}
    \item[] Justification: Code and benchmark would be released after submission and review.
    \item[] Guidelines:
    \begin{itemize}
        \item The answer \answerNA{} means that paper does not include experiments requiring code.
        \item Please see the NeurIPS code and data submission guidelines (\url{https://neurips.cc/public/guides/CodeSubmissionPolicy}) for more details.
        \item While we encourage the release of code and data, we understand that this might not be possible, so \answerNo{} is an acceptable answer. Papers cannot be rejected simply for not including code, unless this is central to the contribution (e.g., for a new open-source benchmark).
        \item The instructions should contain the exact command and environment needed to run to reproduce the results. See the NeurIPS code and data submission guidelines (\url{https://neurips.cc/public/guides/CodeSubmissionPolicy}) for more details.
        \item The authors should provide instructions on data access and preparation, including how to access the raw data, preprocessed data, intermediate data, and generated data, etc.
        \item The authors should provide scripts to reproduce all experimental results for the new proposed method and baselines. If only a subset of experiments are reproducible, they should state which ones are omitted from the script and why.
        \item At submission time, to preserve anonymity, the authors should release anonymized versions (if applicable).
        \item Providing as much information as possible in supplemental material (appended to the paper) is recommended, but including URLs to data and code is permitted.
    \end{itemize}

\item {\bf Experimental setting/details}
    \item[] Question: Does the paper specify all the training and test details (e.g., data splits, hyperparameters, how they were chosen, type of optimizer) necessary to understand the results?
    \item[] Answer: \answerYes{} % Replace by \answerYes{}, \answerNo{}, or \answerNA{}.
    \item[] Justification: Details of training and testing are specified in Section 4.
    \item[] Guidelines:
    \begin{itemize}
        \item The answer \answerNA{} means that the paper does not include experiments.
        \item The experimental setting should be presented in the core of the paper to a level of detail that is necessary to appreciate the results and make sense of them.
        \item The full details can be provided either with the code, in appendix, or as supplemental material.
    \end{itemize}

\item {\bf Experiment statistical significance}
    \item[] Question: Does the paper report error bars suitably and correctly defined or other appropriate information about the statistical significance of the experiments?
    \item[] Answer: \answerNo{}
    \item[] Justification: Experiments in this paper are conducted in multiple iterations, yielding stable
results.
    \item[] Guidelines:
    \begin{itemize}
        \item The answer \answerNA{} means that the paper does not include experiments.
        \item The authors should answer \answerYes{} if the results are accompanied by error bars, confidence intervals, or statistical significance tests, at least for the experiments that support the main claims of the paper.
        \item The factors of variability that the error bars are capturing should be clearly stated (for example, train/test split, initialization, random drawing of some parameter, or overall run with given experimental conditions).
        \item The method for calculating the error bars should be explained (closed form formula, call to a library function, bootstrap, etc.)
        \item The assumptions made should be given (e.g., Normally distributed errors).
        \item It should be clear whether the error bar is the standard deviation or the standard error of the mean.
        \item It is OK to report 1-sigma error bars, but one should state it. The authors should preferably report a 2-sigma error bar than state that they have a 96\% CI, if the hypothesis of Normality of errors is not verified.
        \item For asymmetric distributions, the authors should be careful not to show in tables or figures symmetric error bars that would yield results that are out of range (e.g., negative error rates).
        \item If error bars are reported in tables or plots, the authors should explain in the text how they were calculated and reference the corresponding figures or tables in the text.
    \end{itemize}

\item {\bf Experiments compute resources}
    \item[] Question: For each experiment, does the paper provide sufficient information on the computer resources (type of compute workers, memory, time of execution) needed to reproduce the experiments?
    \item[] Answer: \answerYes{} % Replace by \answerYes{}, \answerNo{}, or \answerNA{}.
    \item[] Justification: The computational cost and resources are specified in Section 5.

    \item[] Guidelines:
    \begin{itemize}
        \item The answer \answerNA{} means that the paper does not include experiments.
        \item The paper should indicate the type of compute workers CPU or GPU, internal cluster, or cloud provider, including relevant memory and storage.
        \item The paper should provide the amount of compute required for each of the individual experimental runs as well as estimate the total compute. 
        \item The paper should disclose whether the full research project required more compute than the experiments reported in the paper (e.g., preliminary or failed experiments that didn't make it into the paper). 
    \end{itemize}
    
\item {\bf Code of ethics}
    \item[] Question: Does the research conducted in the paper conform, in every respect, with the NeurIPS Code of Ethics \url{https://neurips.cc/public/EthicsGuidelines}?
    \item[] Answer: \answerYes{} % Replace by \answerYes{}, \answerNo{}, or \answerNA{}.
    \item[] Justification:  The research conducted in this paper conforms in every respect with the
NeurIPS Code of Ethics.
    \item[] Guidelines:
    \begin{itemize}
        \item The answer \answerNA{} means that the authors have not reviewed the NeurIPS Code of Ethics.
        \item If the authors answer \answerNo, they should explain the special circumstances that require a deviation from the Code of Ethics.
        \item The authors should make sure to preserve anonymity (e.g., if there is a special consideration due to laws or regulations in their jurisdiction).
    \end{itemize}

\item {\bf Broader impacts}
    \item[] Question: Does the paper discuss both potential positive societal impacts and negative societal impacts of the work performed?
    \item[] Answer: \answerYes{} % Replace by \answerYes{}, \answerNo{}, or \answerNA{}.
    \item[] Justification:  The Broader Impacts are discussed in Section 5.
    \item[] Guidelines:
    \begin{itemize}
        \item The answer \answerNA{} means that there is no societal impact of the work performed.
        \item If the authors answer \answerNA{} or \answerNo, they should explain why their work has no societal impact or why the paper does not address societal impact.
        \item Examples of negative societal impacts include potential malicious or unintended uses (e.g., disinformation, generating fake profiles, surveillance), fairness considerations (e.g., deployment of technologies that could make decisions that unfairly impact specific groups), privacy considerations, and security considerations.
        \item The conference expects that many papers will be foundational research and not tied to particular applications, let alone deployments. However, if there is a direct path to any negative applications, the authors should point it out. For example, it is legitimate to point out that an improvement in the quality of generative models could be used to generate Deepfakes for disinformation. On the other hand, it is not needed to point out that a generic algorithm for optimizing neural networks could enable people to train models that generate Deepfakes faster.
        \item The authors should consider possible harms that could arise when the technology is being used as intended and functioning correctly, harms that could arise when the technology is being used as intended but gives incorrect results, and harms following from (intentional or unintentional) misuse of the technology.
        \item If there are negative societal impacts, the authors could also discuss possible mitigation strategies (e.g., gated release of models, providing defenses in addition to attacks, mechanisms for monitoring misuse, mechanisms to monitor how a system learns from feedback over time, improving the efficiency and accessibility of ML).
    \end{itemize}
    
\item {\bf Safeguards}
    \item[] Question: Does the paper describe safeguards that have been put in place for responsible release of data or models that have a high risk for misuse (e.g., pre-trained language models, image generators, or scraped datasets)?
    \item[] Answer: \answerNo{} % Replace by \answerYes{}, \answerNo{}, or \answerNA{}.
    \item[] Justification: Our work does not pose risks requiring such safeguards.
    \item[] Guidelines:
    \begin{itemize}
        \item The answer \answerNA{} means that the paper poses no such risks.
        \item Released models that have a high risk for misuse or dual-use should be released with necessary safeguards to allow for controlled use of the model, for example by requiring that users adhere to usage guidelines or restrictions to access the model or implementing safety filters. 
        \item Datasets that have been scraped from the Internet could pose safety risks. The authors should describe how they avoided releasing unsafe images.
        \item We recognize that providing effective safeguards is challenging, and many papers do not require this, but we encourage authors to take this into account and make a best faith effort.
    \end{itemize}

\item {\bf Licenses for existing assets}
    \item[] Question: Are the creators or original owners of assets (e.g., code, data, models), used in the paper, properly credited and are the license and terms of use explicitly mentioned and properly respected?
    \item[] Answer: \answerYes{} % Replace by \answerYes{}, \answerNo{}, or \answerNA{}.
    \item[] Justification: All models and datasets used in this paper have been properly credited.
    \item[] Guidelines:
    \begin{itemize}
        \item The answer \answerNA{} means that the paper does not use existing assets.
        \item The authors should cite the original paper that produced the code package or dataset.
        \item The authors should state which version of the asset is used and, if possible, include a URL.
        \item The name of the license (e.g., CC-BY 4.0) should be included for each asset.
        \item For scraped data from a particular source (e.g., website), the copyright and terms of service of that source should be provided.
        \item If assets are released, the license, copyright information, and terms of use in the package should be provided. For popular datasets, \url{paperswithcode.com/datasets} has curated licenses for some datasets. Their licensing guide can help determine the license of a dataset.
        \item For existing datasets that are re-packaged, both the original license and the license of the derived asset (if it has changed) should be provided.
        \item If this information is not available online, the authors are encouraged to reach out to the asset's creators.
    \end{itemize}

\item {\bf New assets}
    \item[] Question: Are new assets introduced in the paper well documented and is the documentation provided alongside the assets?
    \item[] Answer: \answerNo{} % Replace by \answerYes{}, \answerNo{}, or \answerNA{}.
    \item[] Justification: The paper does not release new assets.
    \item[] Guidelines:
    \begin{itemize}
        \item The answer \answerNA{} means that the paper does not release new assets.
        \item Researchers should communicate the details of the dataset\slash code\slash model as part of their submissions via structured templates. This includes details about training, license, limitations, etc. 
        \item The paper should discuss whether and how consent was obtained from people whose asset is used.
        \item At submission time, remember to anonymize your assets (if applicable). You can either create an anonymized URL or include an anonymized zip file.
    \end{itemize}

\item {\bf Crowdsourcing and research with human subjects}
    \item[] Question: For crowdsourcing experiments and research with human subjects, does the paper include the full text of instructions given to participants and screenshots, if applicable, as well as details about compensation (if any)? 
    \item[] Answer: \answerNA{} % Replace by \answerYes{}, \answerNo{}, or \answerNA{}.
    \item[] Justification:  The paper does not involve crowdsourcing nor research with human subjects.
    \item[] Guidelines:
    \begin{itemize}
        \item The answer \answerNA{} means that the paper does not involve crowdsourcing nor research with human subjects.
        \item Including this information in the supplemental material is fine, but if the main contribution of the paper involves human subjects, then as much detail as possible should be included in the main paper. 
        \item According to the NeurIPS Code of Ethics, workers involved in data collection, curation, or other labor should be paid at least the minimum wage in the country of the data collector. 
    \end{itemize}

\item {\bf Institutional review board (IRB) approvals or equivalent for research with human subjects}
    \item[] Question: Does the paper describe potential risks incurred by study participants, whether such risks were disclosed to the subjects, and whether Institutional Review Board (IRB) approvals (or an equivalent approval/review based on the requirements of your country or institution) were obtained?
    \item[] Answer: \answerNA{} % Replace by \answerYes{}, \answerNo{}, or \answerNA{}.
    \item[] Justification: The paper does not involve crowdsourcing nor research with human subjects.
    \item[] Guidelines:
    \begin{itemize}
        \item The answer \answerNA{} means that the paper does not involve crowdsourcing nor research with human subjects.
        \item Depending on the country in which research is conducted, IRB approval (or equivalent) may be required for any human subjects research. If you obtained IRB approval, you should clearly state this in the paper. 
        \item We recognize that the procedures for this may vary significantly between institutions and locations, and we expect authors to adhere to the NeurIPS Code of Ethics and the guidelines for their institution. 
        \item For initial submissions, do not include any information that would break anonymity (if applicable), such as the institution conducting the review.
    \end{itemize}

\item {\bf Declaration of LLM usage}
    \item[] Question: Does the paper describe the usage of LLMs if it is an important, original, or non-standard component of the core methods in this research? Note that if the LLM is used only for writing, editing, or formatting purposes and does \emph{not} impact the core methodology, scientific rigor, or originality of the research, declaration is not required.
    %this research? 
    \item[] Answer: \answerNA{} % Replace by \answerYes{}, \answerNo{}, or \answerNA{}.
    \item[] Justification: LLM is only used for writing.
    \item[] Guidelines:
    \begin{itemize}
        \item The answer \answerNA{} means that the core method development in this research does not involve LLMs as any important, original, or non-standard components.
        \item Please refer to our LLM policy in the NeurIPS handbook for what should or should not be described.
    \end{itemize}

\end{enumerate}